\documentclass[sigconf]{acmart}
\renewcommand\footnotetextcopyrightpermission[1]{} 

\setcopyright{rightsretained}
\acmConference[SIGIR eCom'22]{ACM SIGIR Workshop on eCommerce}{July 15, 2022}{Madrid, Spain}
\acmYear{2022}
\copyrightyear{2022}
\makeatletter
\renewcommand\@formatdoi[1]{\ignorespaces}
\makeatother
\acmISBN{}

\usepackage{amsmath}
\usepackage{graphicx}
\usepackage{float}
\usepackage[caption = false]{subfig}
\usepackage{longtable}
\usepackage{booktabs}
\usepackage{multirow}
\usepackage{siunitx}
\usepackage{algorithm}
\usepackage[noend]{algpseudocode}
\usepackage{mathpazo}

\AtBeginDocument{%
  \providecommand\BibTeX{{%
    \normalfont B\kern-0.5em{\scshape i\kern-0.25em b}\kern-0.8em\TeX}}}

\begin{document}

\title{Vernacular Search Query Translation with Unsupervised Domain Adaptation}
\author{Mandar Kulkarni, Nikesh Garera}
\affiliation{%
 \institution{Flipkart Data Science}
 \country{India}}
\email{(mandar.kulkarni, nikesh.garera)@flipkart.com}



\begin{abstract}

With the democratization of e-commerce platforms, an increasingly diversified user base is opting to shop online. To provide a comfortable and reliable shopping experience, it's important to enable users to interact with the platform in the language of their choice. An accurate query translation is essential for Cross-Lingual Information Retrieval (CLIR) with vernacular queries. Due to internet-scale operations, e-commerce platforms get millions of search queries every day. However, creating a parallel training set to train an in-domain translation model is cumbersome. 
This paper proposes an unsupervised domain adaptation approach to translate search queries without using any parallel corpus. We use an open-domain translation model (trained on public corpus) and adapt it to the query data using only the monolingual queries from two languages. In addition, fine-tuning with a small labeled set further improves the result. For demonstration, we show results for Hindi to English query translation and use mBART-large-50 model as the baseline to improve upon. Experimental results show that, without using any parallel corpus, we obtain more than 20 BLEU points improvement over the baseline while fine-tuning with a small 50k labeled set provides more than 27 BLEU points improvement over the baseline.

\end{abstract}





\maketitle

\thispagestyle{fancy}
\fancyhf{}
\rhead{Accepted @ SIGIR e-Commerce Workshop 2022}

\section{Introduction}

With the democratization of e-commerce platforms, an increasingly diversified user base is opting to shop online. To provide a comfortable and reliable shopping experience, it's important to enable users to interact with the platform in the language of their choice. To enable Cross-Lingual Information Retrieval (CLIR), vernacular search queries need to be translated. 

In this paper, we propose an unsupervised domain adaptation approach to translate vernacular queries without using any parallel corpus. For demonstration, we show results for Hindi to English query translation. Due to internet-scale operations, e-commerce platforms get millions of search queries every day. Therefore, in-domain unlabeled query data is available in large volumes. However, creating a large parallel training corpus for training the translation model is cumbersome. 
We use an open-domain trainable translation model (trained on a large publicly available corpus) and adapt it for search query translation using only the monolingual queries from both languages. For unsupervised domain adaptation, we experiment with unsupervised NMT techniques such as cross-domain training, denoising auto-encoder, and adversarial updates \cite{lample2018unsupervised}\cite{artetxe2018unsupervised}. 


As the open-domain NMT model, we use mBART-large-50-many-to-many-mmt as the baseline \cite{tang2020multilingual} and further adapt it for query translation. For ease of reading, from now on, we will refer to this model as mBART-50. Experimental results show that we get more than 20 BLEU points improvements over the baseline with domain adaptation without using any parallel corpus. In addition, finetuning the domain-adapted model on a small set of 50k labeled queries provides more than 27 BLEU points improvements over the baseline.
Though we demonstrate the effectiveness of the approach for query translation, the proposed approach is generic in nature and can potentially be used for NMT with unsupervised domain adaptation for different domains. 

The main contributions of the paper is as follows.
\begin{itemize}
  \item Proposed an approach to use unsupervised NMT techniques \cite{lample2018unsupervised}\cite{artetxe2018unsupervised} with pre-trained NMT models \cite{tang2020multilingual} for unsupervised domain adaptation  
  
\end{itemize}

\section{Related works}


Unsupervised NMT methods have been experimented with in the past. Alexis et al. \cite{DBLP:journals/corr/abs-1710-04087} 
proposed an approach for unsupervised NMT where they build a bilingual dictionary between two languages by aligning monolingual word embedding spaces in an unsupervised way.
Lample et al. \cite{lample2018unsupervised} trained a bi-LSTM  NMT model with iterative unsupervised techniques such as cross-domain training, denoising auto-encoding, and adversarial alignment of the source and target latent spaces. They use word-level translation \cite{DBLP:journals/corr/abs-1710-04087} as an initial step for the training.   
Artetxe et al. \cite{artetxe2018unsupervised} use on-the-fly-backtransaltion along with a neural architecture comprising of a shared encoder and language-specific decoder to perform unsupervised NMT. We experiment with the publicly available mBART-50 transformer model in our work. Self-supervised text reconstructions from the synthetically created noisy text have proved to be a good pre-training objective. Such denoising auto-encoder training has shown improvement for monolingual \cite{lewis2019bart} as well as multilingual \cite{liu2020multilingual} downstream tasks. Specifically, the multilingual mBART model has demonstrated promising results for low resource scenarios. Tang et al. \cite{tang2020multilingual} have further extended this premise specifically for language translation. With appropriate encoder and decoder settings, mBART-50 model supports translation between 50 different languages. 

Domain adaptation for NMT has been studied in literature \cite{saunders2021domain}. Yao et al. \cite{Yao2020DomainTB} propose mixed attention BERT-based translation refinement approach for domain adaptation for query translation. Introducing adapter layers in the pre-trained models has shown promising results with new domains, and language pairs \cite{stickland2021multilingual}.

\section{Proposed approach}
\pagestyle{plain}

We explain an approach to adapting an open-domain NMT model trained on a large public corpus to perform a translation with query data without using any parallel corpus. 
The following sections describe the details of the approach.

\subsection{Data preparation}

For model training, we use an in-house query dataset.
We collected a dataset of 5.06M unique English queries and 5.45M unique Hindi queries. The Hindi queries are detected based on character unicode ranges i.e. any query which contains atleast one Hindi character is considered as Hindi query. The query data is fetched from the database with standard SQL queries. A spell correction is applied to raw English queries before using them for training. Note that the training dataset does not have any parallel data.

We use 20k manually tagged Hindi queries as the test set for model evaluation. To terminate the training, we use 100k unlabeled English queries as the validation set. 


\subsection{Model details}

We use the mBART-50 model \cite{tang2020multilingual} which is trained on publicly available data such as WMT, IWSLT, WAT, TED. The model can translate between any pair of supported set of 50 languages. There are three variants of the mBART-50 model available: Many-to-one, one-to-Many, and Many-to-Many. We use the Many-to-Many variant to facilitate cross-language training as explained in section \ref{sec:CrossLT}. The model has a vocab size of 250k and approx. 610M trainable parameters. It has 12 layers in the encoder and the decoder.
Given any input language (Hindi/English), the mBART-50 model can be forced to generate text from the intended language by setting the 'forced\_bos\_token\_id' token to "en\_XX"/ "hi\_IN".

\subsection{Model Training}

For adapting the model to query data, we experiment with the following unsupervised NMT objectives: Denoising autoencoder, cross-language training, and adversarial updates \cite{lample2018unsupervised} \cite{artetxe2018unsupervised}.

\subsubsection{\textbf{Denoising AutoEncoder (DenoiseAE)}}

For this objective, the model is trained to reconstruct the text from the synthetically created noisy version of it. For mBART, since the encoder and the decoder are shared between all languages, we use this training update with Hindi as well as English queries. We use cross-entropy loss as the training objective. We experimented with the following noise types.

\begin{itemize}
  \item \textbf{Mask}: Since search queries are short in length, we randomly mask a single word from the Hindi/English query and train the model to reconstruct the entire query. A randomly selected word is replaced with the '[MASK]' token. Note that a word may correspond to one or more consecutive tokens.
  
  \item \textbf{DropChar}: We randomly drop a character (from the middle of the word) for 30-50\% of the words in the Hindi/English query and train the model to reconstruct the query.  
  
  \item \textbf{Shuffle}: We randomly permute the order of the words in the Hindi/English query and train the model to de-noise it.

\end{itemize}

\begin{table*}[!t]
\centering
  \begin{tabular}{ll}
    \toprule
    \multirow{1}{*}{Setting} &  
     \multirow{1}{*}{BLEU} \\
       
      \midrule
      mBART & 25.7 \\
      mBART with One-time Backtranslation & 36.5 \\
      mBART with CrossLT & 44.8 \\
      mBART with CrossLT + Adv & 42.2 \\
      
      mBART with CrossLT + DenoiseAE (Shuffle) & 38.8\\
      mBART with CrossLT + DenoiseAE (Mask) & 44.1\\
      mBART with CrossLT + DenoiseAE (DropChar) & \textbf{46.1}\\
    \bottomrule
  \end{tabular}
  
  \caption{\label{table:res0} Results on the test set without using any parallel corpus.}
  
\end{table*}

\subsubsection{\textbf{Cross Language Training (CrossLT)}}
\label{sec:CrossLT}

We use a two-stage Cross-Language Training (CrossLT) approach to mimic the translation task with unlabeled search queries. In the first stage, given an input search query in Hindi/English, we use the mBART model in the inference mode with greedy decoding to translate it to English/Hindi. We set the \newline 'forced\_bos\_token\_id' token to an appropriate language. In the second stage, the generated synthetic translation is used to predict the original query using the teacher forcing strategy. Note that, the synthetic translation from the first stage is now used as the input and the original query text is treated as the target. We use cross-entropy loss for this update. The approach is similar to on-the-fly back-translation, where instead of using an independent model, the mBART-50 itself is used to back-translate a query batch at a time. As the training progresses, the model would produce better synthetic query pairs through back-translation, which helps to further improve the model in the subsequent iterations. 
For this objective, since the decoder need to generate the text from both languages, only a Many-to-Many setting of the mBART-50 model is appropriate. This objective is found to be crucial for the success of unsupervised NMT \cite{lample2018unsupervised}. Intuitively, this objective will fail if the initial model does not produce decent translations. This is because, if the initial model produces very noisy translations, the teacher forcing update in the second stage will mostly be done with noise as the input, and the conditional generation model will act as the language model. 
To avoid the possibility of this de-generate solution, authors of \cite{lample2018unsupervised} 
use word-level translations as the initial model. In our case, we observe that mBART-50 model already provides decent translations for in-domain query data, which are then significantly improved with the CrossLT update.

\begin{figure*} 
\centering

\begin{tabular}{c}

\includegraphics[width=\linewidth, height = 150pt]{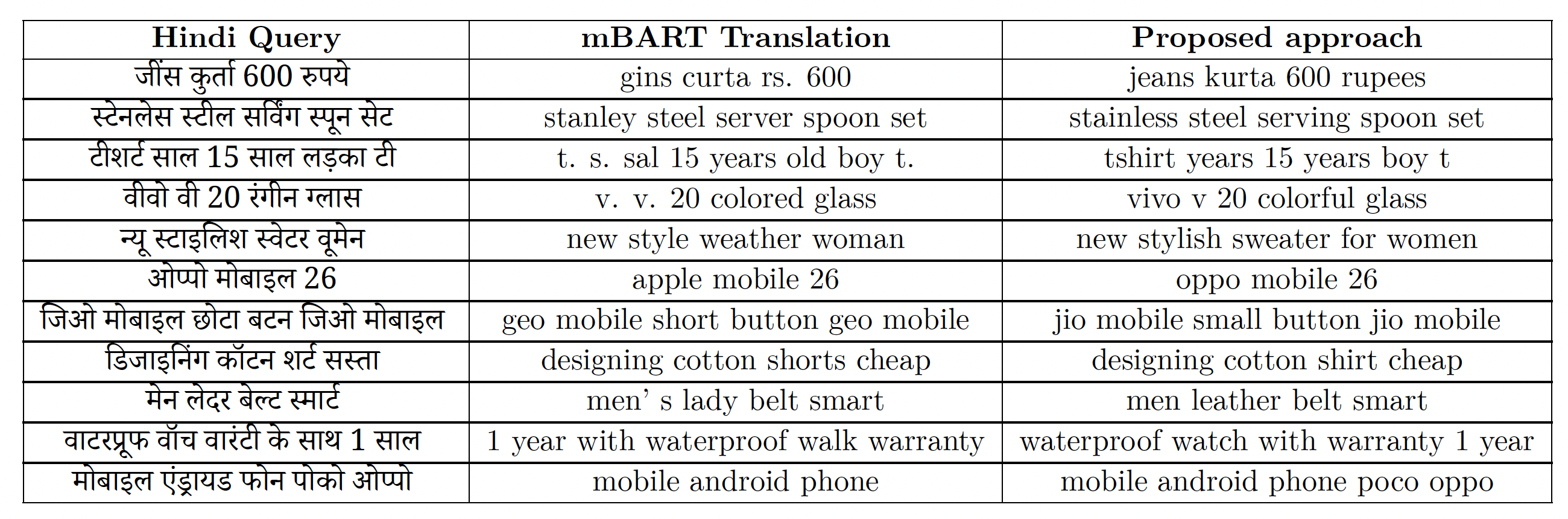}

\end{tabular}
\caption{\label{fig:inpt} Hindi search query translation results: mBART translation indicates result obtained with open-domain mBART-50 NMT model. Proposed approach column indicates translation results (using the best setting) obtained with unsupervised domain adaptation i.e. NMT model trained without using any parallel corpus.}
\end{figure*}

\subsubsection{\textbf{Adversarial update (Adv)}}

Aligning word-level features from the encoder with adversarial updates has provided better accuracies with unsupervised NMT \cite{lample2018unsupervised}. We experiment with the adversarial update to align encoder token-level embeddings from both languages. We use mBART's encoder as a feature generator and use a 2 layer fully connected network as the discriminator. For the discriminator update, the model is trained to identify the language of the input queries. Instead of hard labels, we use soft labels to train the discriminator. To update the generator, for the features of the English queries, we use Hindi as the language label to confuse the discriminator about the language of the input queries. 

For model training, we follow a sequential update strategy as used in \cite{artetxe2018unsupervised}. DenoiseAE, CrossLT, and Adv updates are done sequentially based on the experiment setting. 
For the model update for each objective, a batch of monolingual queries is sampled randomly. For the DenoiseAE update, for each batch, a type of noise is chosen randomly, and the model is updated using Hindi and English query data. 
Similarly, for CrossLT, the model updates for Hindi and English queries are executed subsequently.  
Since unlabeled queries are used for training, we use 100k English queries as the validation set to halt the training process. During validation evaluation, all queries in the validation set are translated to Hindi and the resulting outputs are translated back to English. We then calculate BLEU scores between original English queries and their reconstructions. Training is terminated when the BLEU score does not improve for 3 consecutive evaluations. The validation data is evaluated after 5k updates for each objective. 
The learning rate is set to 5e-6, while the batch size is set to 16. For training, we use label smoothing, where the smoothing parameter is set to 0.1. During the inference, we use beam search decoding with a beam size of 3. 

\begin{figure*} [!t]
\centering
\begin{tabular}{c c c}
\includegraphics[width=150pt, height = 100pt]{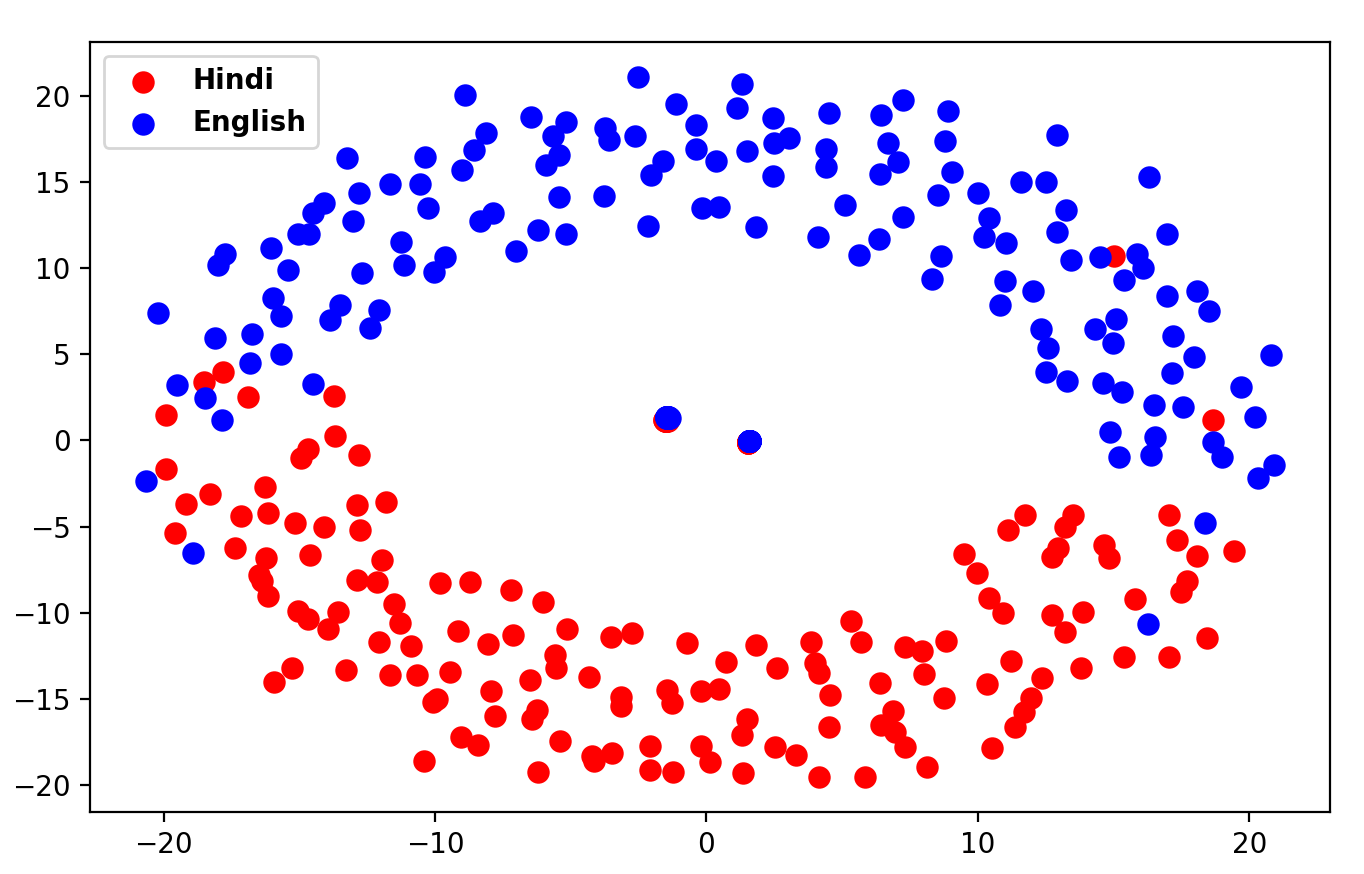}&
\includegraphics[width=150pt, height = 100pt]{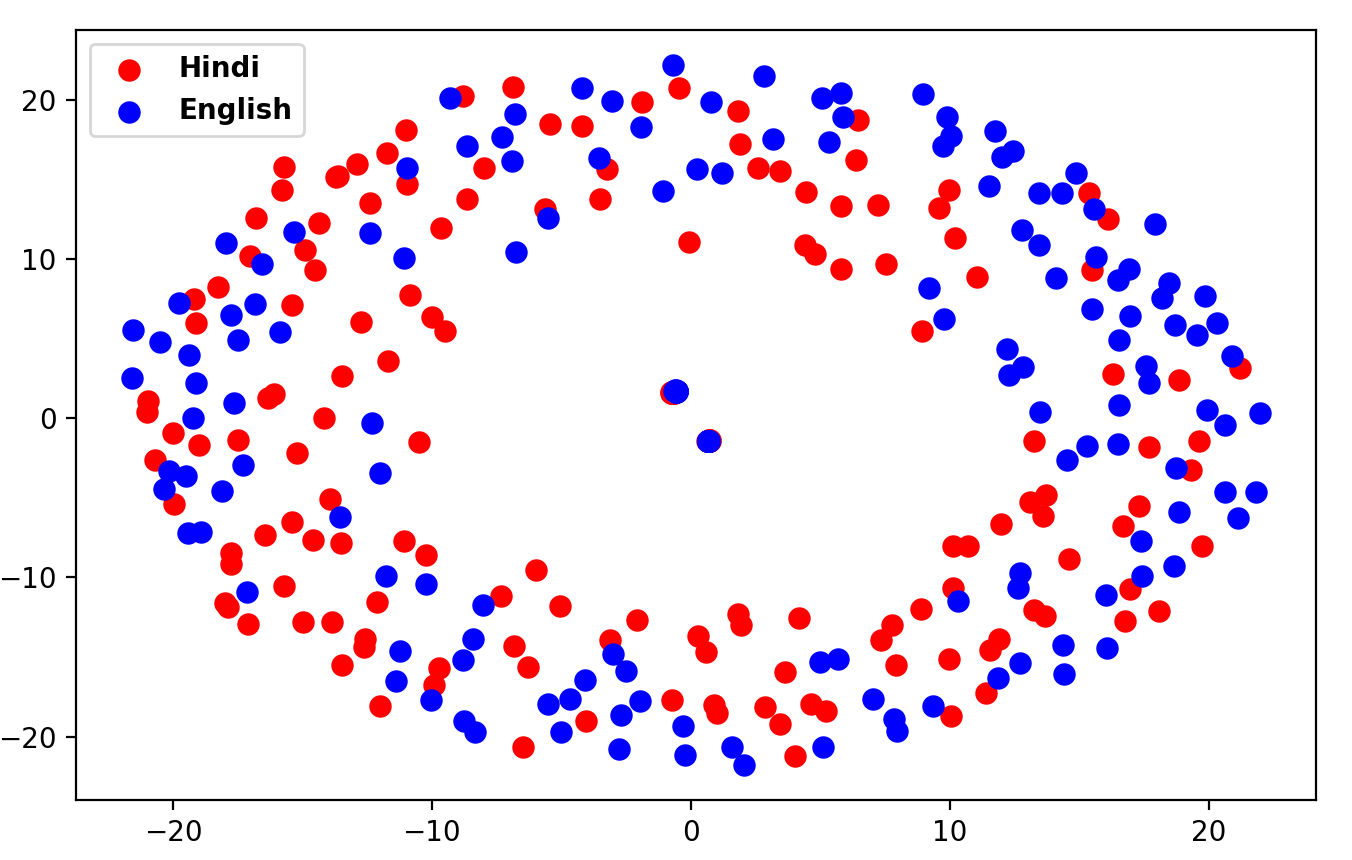}&
\includegraphics[width=150pt, height=100pt]{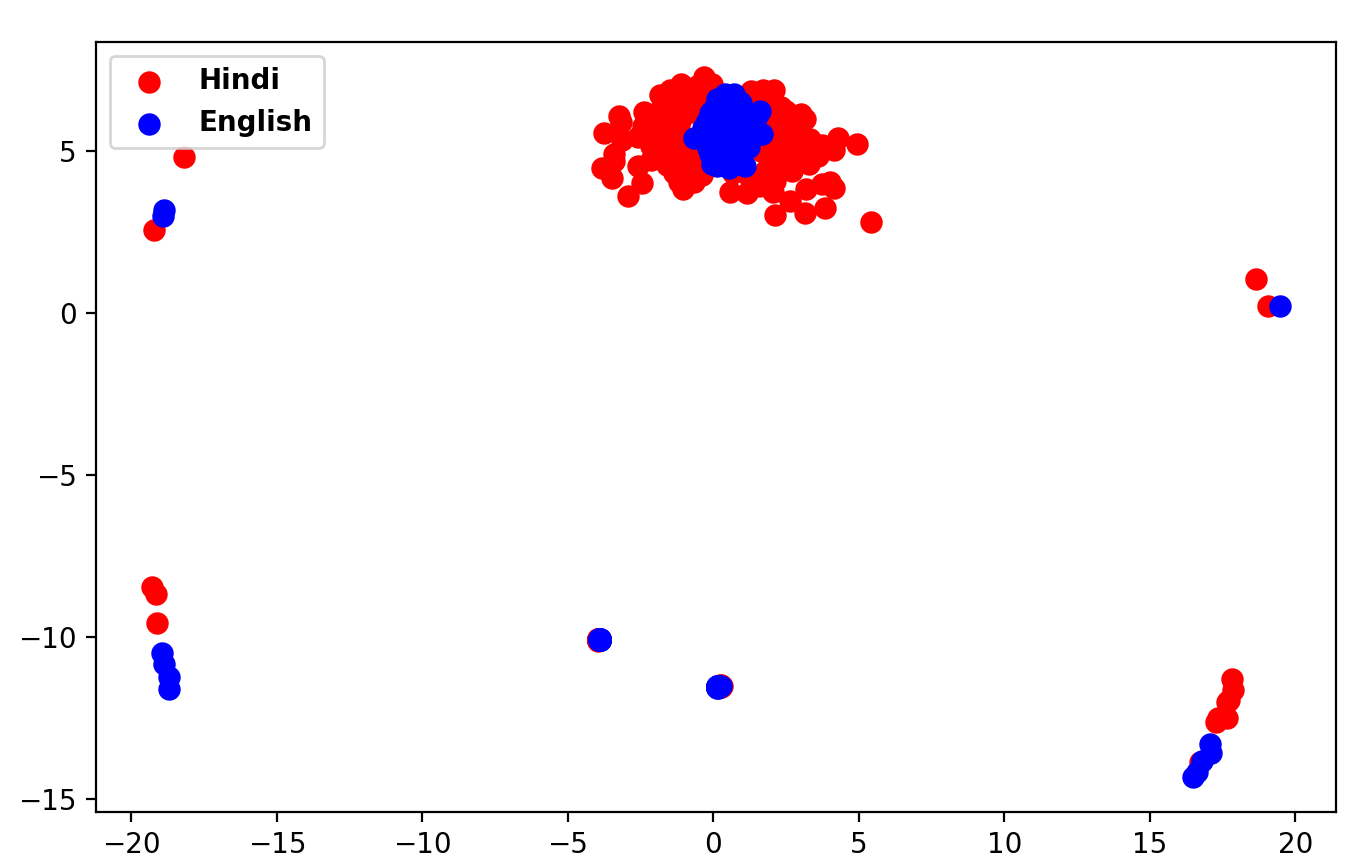}\\(a)&(b)&(c)\\

\end{tabular}
\caption{\label{fig:inpt116} Effect of adversarial update on encoder features. (a) Baseline mBART-50 model, (b) features after CrossLT and adversarial updates, (c) features after 2k only adversarial updates,. Red indicates Hindi features while Blue indicates English features.}
\end{figure*}

We use 20k manually tagged Hindi queries as the test set for model evaluation. Table \ref{table:res0} shows the result of different model settings on the test set. All the BLEU scores are computed using SacreBLEU \cite{post2018clarity}. The baseline result indicates the BLEU score of mBART-50 model on the test set. Interestingly, the model trained only with the CrossLT update provides more than 19 points BLEU points improvement over the Baseline. This is surprising because, with the CrossLT update, the model is trained just by feeding its translation outputs as the input to reconstruct the original text. 
CrossLT combined with adversarial updates provides relatively less accuracy. We further analyze the result with adversarial updates, and details are given in the  section \ref{sec:adv}. For CrossLT + DenoiseAE setting, DropChar provides improvement over only CrossLT update. Results with query word shuffle are relatively inferior compared to the CrossLT update. Possibly due to the lack of grammar in the majority of search queries, the mBART model struggles to reconstruct the original word order. Masking works fairly well but does not provide an improvement over only CrossLT updates. CrossLT, along with DropChar denoise, provides more than 20 BLEU points improvement over the Baseline, indicating the efficacy of the proposed approach. Figure \ref{fig:inpt} shows the comparison result for query translation. Note that the proposed model is better at fixing spelling errors and word mismatch issues. 

To verify the effectiveness of iterative cross-lingual back-translation, we compare the result with one-time back-translation. We back-translate the 5.06M English query dataset to Hindi using the mBART-50 model and then train a forward model for Hindi to English translation. The second row in Table \ref{table:res0} shows the result of the experiment. The result indicates that iterative back-translation is a more effective approach for domain adaptation.

\section{Finetuning with labeled set}

If a small manually labeled set of queries is available, it can help to improve the translation results further.  
We finetune the domain adapted (CrossLT + DenoiseAE-DropChar) model on the set of 10k and 50k labeled queries where 10\% of the queries are used for validation. The learning rate is set to 1e-5. Table \ref{tab:res10} shows the result. Finetuning with manually labeled corpus provides significant improvements, whereas training with only 50k labeled samples provides more than 27 BLEU points improvement over the Baseline.  


\begin{table}[h]
    \centering
    \captionsetup{justification=centering, margin=5mm}
    \begin{tabular}{|c|c|}
        
        \hline
        \textbf{Model} & \textbf{BLEU}\\
        \hline
        CrossLT + Denoise-DropChar & 46.1 \\
        \hline
        Fintuning with 10k & 50.5 \\
        \hline
        Fintuning with 50k & \textbf{53.6} \\
        \hline

        \end{tabular}
    \vspace{2mm}
    \caption{Results with fine-tuning}
    \label{tab:res10}
\end{table}

\section{Effect of adversarial update}
\label{sec:adv}

We analyzed the effect of adversarial updates on the mBART-50's encoder feature representations. For the same set of unlabeled queries from Hindi and English and with different model settings, we extract the token-level features from the encoder and project them onto 2D space using MDS dimension reduction for visualization. 
Figure \ref{fig:inpt116} shows the resultant plots. Hindi and English encoder features are indicated in red and blue, respectively. Figure \ref{fig:inpt116} (a) shows the result for the Baseline mBART-50 model while Figure \ref{fig:inpt116} (b) shows encoder features for the model trained with CrossLT and adversarial updates. Note that the mBART-50 token embeddings for Hindi and English queries show minimal overlap, while a more accurate domain-adapted model (with CrossLT + adversarial updates) shows a high degree of overlap. This may indicate that, for a good accuracy multilingual translation model, it is crucial to have aligned encoder embeddings for different input languages.

Figure \ref{fig:inpt116} (c) shows the features after 2k adversarial updates without CrossLT update. An adversarial update maps two feature spaces close to each other; however, it does not preserve the inter-token similarity.
In fact, with only adversarial updates, model training failed and did not give any improvement over the Baseline. Hence, good accuracy with CrossLT + adversarial updates can be attributed to the CrossLT training objective.
Lample et al.\cite{lample2018unsupervised} have shown that adversarial updates work well with word-level features; however, in our case, aligning the token-level features along with the CrossLT update did not give improvement over only the CrossLT update. This could be because the sub-word tokenizer trained on a specific domain is being applied to a new domain where individual tokens may not necessarily have a semantic meaning.

\section{Conclusion}

In this paper, we proposed an unsupervised domain adaptation approach for adapting an open-domain mBART-50 translation model for e-commerce query translation. We experimented with different training objectives and found that the Cross-Domain Training combined with Denoising Auto-Encoder provided the most prominent improvement over the baseline mBART-50 model. Additionally, finetuning with a small manually labeled set provided further accuracy improvements. Experimental results demonstrated the efficacy of the proposed approach.

\bibliographystyle{ACM-Reference-Format}
\bibliography{sample-base}

\newpage



\end{document}